\documentclass[conference]{IEEEtran}

  	\usepackage[pdftex]{graphicx}
  	\graphicspath{{../pdf/}{../jpeg/}}
	\DeclareGraphicsExtensions{.pdf,.jpeg,.png}

	\usepackage[font=small]{caption}
	\usepackage[cmex10]{amsmath}
	\usepackage{mathabx}
	\usepackage{array}
	\usepackage{mdwmath}
	\usepackage{mdwtab}
	\usepackage{eqparbox}
	\usepackage{url}
	\usepackage{algorithm}
    \usepackage[noend]{algpseudocode}
    \usepackage[dvipsnames]{xcolor}
    
    \usepackage{subcaption}
    \usepackage{multicol}
    \usepackage{fancyhdr}
    \usepackage{lastpage}
    \usepackage{xcolor}

\renewcommand\tablename{Table}

\begin{document}

\title{\LARGE Comparative Analysis of Agent-Oriented Task Assignment and Path Planning Algorithms Applied to Drone Swarms}


 \author{\authorblockN{Rohith Gandhi Ganesan\authorrefmark{1}\authorrefmark{2}, Samantha Kappagoda\authorrefmark{1}, Giuseppe Loianno\authorrefmark{1}\authorrefmark{2}, David K. A. Mordecai\authorrefmark{1}}
 \authorblockA{\authorrefmark{1}RiskEcon\textsuperscript{®} Lab for Decision Metrics @ Courant Institute of Mathematical Sciences}
 \authorblockA{\authorrefmark{2}Agile Robotics and Perception Lab (ARPL) @ Tandon School of Engineering}
 \authorblockA{New York University, New York, NY 10011 \\ \{rgg296, loiannog\} @nyu.edu, \{kappagoda, mordecai\} @cims.nyu.edu}
 }

\maketitle

\begin{abstract}
Autonomous drone swarms are a burgeoning technology with significant applications in the field of mapping, inspection, transportation and monitoring. To complete a task, each drone has to accomplish a sub-goal within the context of the overall task at hand and navigate through the environment by avoiding collision with obstacles and with other agents in the environment. In this work, we choose the task of optimal coverage of an environment with drone swarms where the global knowledge of the goal states and its positions are known but not of the obstacles. The drones have to choose the Points of Interest (PoI) present in the environment to visit, along with the order to be visited to ensure fast coverage. We model this task in a simulation and use an agent-oriented approach to solve the problem. We evaluate different policy networks trained with reinforcement learning algorithms based on their effectiveness, i.e. time taken to map the area and efficiency, i.e. computational requirements. We couple the task assignment with path planning in an unique way for performing collision avoidance during navigation and compare a grid-based global planning algorithm, i.e. Wavefront and a gradient-based local planning algorithm, i.e. Potential Field. We also evaluate the Potential Field planning algorithm with different cost functions, propose a method to adaptively modify the velocity of the drone when using the Huber loss function to perform collision avoidance and observe its effect on the trajectory of the drones. We demonstrate our experiments in 2D and 3D simulations. \\
\end{abstract}



\IEEEpeerreviewmaketitle


\section{Introduction}
Swarm robotics has been an active area of research and recent improvements in computational capabilities, deep learning and reinforcement learning methods have led to increased interest in autonomous swarm systems. Using a swarm of robots over a single robot to accomplish a task would reduce the time taken to accomplish it, increase the robustness, flexibility and scalability of the system \cite{10.3389/frobt.2020.00018}. Drone swarms, in particular, have significant applications as they can be used to map an unknown area \cite{ozaslan_autonomous_2017, doi:10.1177/095965180321700205, albani_dynamic_2018}, inspect objects which are dangerous or inaccessible to humans \cite{yoshida_vertical_2014, doi:10.1002/rob.20390}, transport object between locations \cite{6329450, 7589023, mellinger_cooperative_2013} and perform aerial monitoring \cite{6386281, saska_swarm_2016}.


 In this work, we consider the task of mapping an unknown area with Points of Interest (PoI) using drone swarms. Points of Interest are locations present in the environment that we would want our drones to reach in order to map them. The drone swarms have to choose the PoI it would visit along with the order in which it would visit them based on the priority level of the points. The drones should also navigate the environment by avoiding collision with obstacles and with each other to reach the PoI. We characterize the first part of the problem, i.e. choosing the order of PoI to visit as a task assignment problem where each drone would choose the order in which it would visit the points to map them and minimize coverage time. The prioritized order is then executed by the path planning algorithm which performs collision avoidance with obstacles and with other drones. Note that the computation of both task assignment and path planning is designed to take place in a decentralized manner, i.e. on the edge device and does not require perfect global knowledge of the environment.

We use Reinforcement Learning (RL) algorithms instead of heuristic based search algorithms (e.g.: Dijkstra, A\textsuperscript{*} Search etc.) to solve the task assignment problem as the search algorithms require exponential time and space complexity. The heuristic based search algorithms do not perform any exploration to gain additional information about the environment. Instead, they only exploit using hard-coded search strategies. They also do not perform an adaptive search strategy for a dynamic environment. Conversely, the \emph{Bellman optimization} method through dynamic programming also suffers from exploding searching state spaces.    
Therefore, using neural networks as function approximators with RL algorithms could enable a method to adaptively and scalably search large state spaces and produce approximately optimal solutions, conditioned upon the relative stationarity of the state space and sufficient computing resources. 
We evaluate the effectiveness of RL based algorithms to exploration problems that involve Unmanned Aerial Vehicles (UAVs). We perform a comparative analysis on several neural network architectures and identify the benefits and drawback of each solution based on its efficiency and effectiveness. We use the policy gradient algorithm with different policy networks including Sequence-to-Sequence model, and Sequence-to-Sequence model with attention, Pointer Net, Transformer networks \cite{yang_survey_2020} and compare their computational requirements and time taken to map the PoI. Furthermore, we couple the aforementioned RL approach to two path planning strategies that does not require perfect knowledge of the environment for obstacle avoidance 
- WaveFront planner \cite{galceran_survey_2013}, which is a \emph{grid-based global planner} and Potential Field planner \cite{zhang_pf}, which is a \emph{gradient-based local planner}. We also study the effect of different cost functions in the Potential Field planner on the trajectory of the drone to allow safety. In order to avoid a completely conservative or radical strategy based on the distance, we propose a method to adaptively change the velocity of the drone by scaling the value of $\delta$ in the Huber loss function. 


\section{Related Work}
\subsection{Task assignment using RL}
The problem of predicting the order of points to be visited is analogous to the \emph{Travelling Salesman Problem (TSP)} where an agent visits all locations with minimal tour length. Given that the TSP is an NP-Hard problem \cite{zhou2003solving}, it intractable to find optimal solutions for large state spaces. Therefore, using neural networks allows the problem to be more tractable by acting as a function approximator and produce an approximately optimal solution. In \cite{vinyals_pointer_2017}, authors propose a new model architecture called Pointer Net and use it to solve the Travelling Salesman Problem through supervised learning. The labels for each input are generated using the \emph{Held-Karp algorithm} \cite{10.1145/321105.321111} for a smaller number of points and for a larger number of points, approximate TSP solvers are used. However, the trained model at best, would attempt to imitate the approximate solvers and produce sub-optimal solutions. \cite{bello_neural_2017} uses the Pointer Net architecture and optimizes the network using the Policy Gradient \cite{sutton_policy_nodate} method and negative tour length as a reward signal. \cite{kool_attention_2019} uses the transformer architecture \cite{vaswani_attention_2017} to solve routing problems including TSP and train the model architecture using Policy Gradient method. \cite{dai_learning_2018-1} proposes a combination of graph neural network and Q-learning to solve the TSP. The above mentioned TSP solution methods are proposed for an environment with a single agent and the prediction order is conditioned upon the tour length and not its priority level. Therefore, in our work, we evaluate the performance of different policy architectures in a \emph{multi-agent} setting for mapping PoI in a simulation. 

\subsection{Path Planning for Drones}
The classical algorithms for performing path planning in robots are Wavefront, Dynamic Window \cite{fox_dynamic_1997}, Potential Field \cite{1087247} and Velocity Obstacle method. \cite{van_den_berg_reciprocal_2008} proposes a variation of the velocity obstacle method used for performing obstacle avoidance in a multi-agent setting. \cite{7337041} applies the Potential Field algorithm to multi-agent Unmanned Aerial Vehicles (UAVs) to avoid static and dynamic obstacles in simulation. In our work, we evaluate the Wavefront and Potential Field algorithm in a multi-agent setting and also study the effects of different cost functions used for the Potential Field algorithm in simulation in 2D and 3D environments. 

\section{Problem Formulation}
We split the problem of performing optimal coverage by drone swarms into two parts - Task Assignment for predicting the PoI to be visited and Path Planning to perform obstacle avoidance. The task assignment problem is described as follows. We are interested in the task of mapping an environment using drone swarms with minimum coverage time. The environment contains certain PoI which must be mapped. Only when all the PoI are mapped is the task considered completed. Some PoI have a higher priority level than others. Therefore, we prefer that the drones visit these prior to visiting PoI with a relatively lower priority. Each drone has to predict the PoI to be visited along with the order in which it would visit them to minimize total steps taken and map all PoI. The path planning algorithm utilizes the predicted order to visit the PoI and creates a trajectory to avoid collision with obstacles and reach the goal points.  

\section{Task Assignment}
In the task assignment problem, the drones predict the PoI to be visited along with the order to be visited such that the total number of time steps to completely map the environment is minimized. The drones use the global information about the location of PoI and independently choose actions that enable them to map the environment as fast as possible. The actions predicted at each time step is obtained from a policy network which is trained using an policy gradient algorithm. 
\subsection{Simulation - 2D}
\begin{figure}
    \centering
    \includegraphics[width=0.8\linewidth]{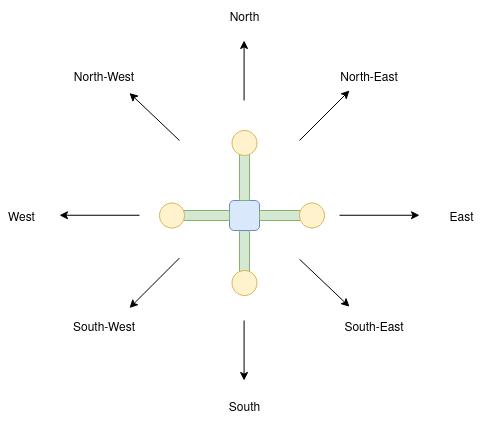}
    \caption{The image showcases 8 possible actions that can be taken by a drone at time step during the task assignment.}
    \label{fig:action_space}
\end{figure}

\begin{figure}
    \centering
    \includegraphics[width=0.8\linewidth]{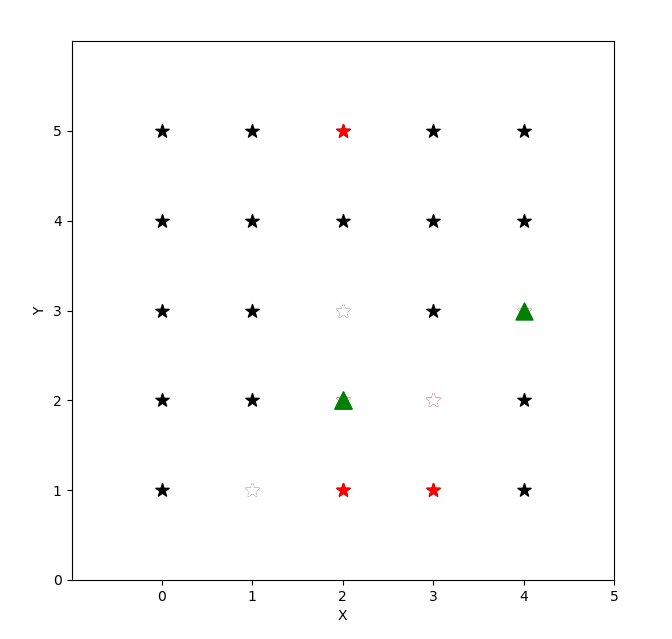}
    \caption{A rendered image from the 2D simulation of $5\times 5$ grid. The green triangles represent the drones, the stars represent the PoI. The red stars have a higher priority level compared to black stars. The white stars show that these points are mapped by the drone}
    \label{fig:sim_sample}
\end{figure}

We model the problem of task assignment in a 2D simulation where the environment is represented as a grid and the drones are spawned at random locations within the environment. The PoI are uniformly distributed within the grid and a certain number of points are randomly sampled and assigned a higher priority level than others. The drones can choose one out of eight possible actions, i.e. 8 possible directions it can move which are North, North-East, East, South-East, South, South-West, West, North-West. A rendered image of the 2D simulation is displayed in Figure \ref{fig:sim_sample} and all possible actions that can be taken by a drone is shown in Figure \ref{fig:action_space}.

\subsection{Reward Function}
The incentive structure of the simulation plays an important role as it models the task we are trying to achieve and it directly affects the performance of the reinforcement learning algorithm.  Equation (\ref{reward}) shows the total reward calculation. Equation (\ref{reward1}) represents the positive reward each drone gets for mapping a point. It obtains a reward of $\alpha$ for mapping a low priority point and a reward of $2\alpha$ for mapping a high priority point, where $\alpha$ is a positive real number. In our experiments we set $\alpha = 5$.

\begin{equation} \label{reward1}
    R_1 = 
    \begin{cases}
    \alpha & \text{if low priority point mapped}\\
    2\alpha & \text{if high priority point mapped}
    \end{cases}
\end{equation}

\begin{equation} \label{reward2}
    R_2 = 
    \begin{cases}
    0 & \text{high priority points mapped}\\
    \min\limits_{i} \| d - x_i \|_2 & \text{otherwise}
    \end{cases}
\end{equation}

\begin{equation} \label{reward}
    \textit{Reward} = R_1 - R_2 - 1
\end{equation}

Equation (\ref{reward2}) represents the penalty the drone receives for delaying the mapping of high priority points. If all high priority points are mapped, the drone does not receive any penalty, else, the drone is penalized by the $L_2$ norm between the drone's location and the closest unmapped high priority point. Equation (\ref{reward}) represents the final reward signal that drones receive. The agent\footnote{Drone and agent is used interchangeably for the rest of the paper.} receives a negative reward of value one at each time step which encourages the policy to map the points faster as it would maximize the cumulative sum of rewards. 

\subsection{Policy Networks}
We evaluate Sequence-to-Sequence (Seq2Seq) model, Sequence-to-Sequence model with attention, Pointer Net and Transformer networks as our policy network and train them using REINFORCE \cite{williams_simple_1992} algorithm.

\subsubsection{Seq2Seq}
\begin{figure}
    \centering
    \includegraphics[width=0.8\linewidth]{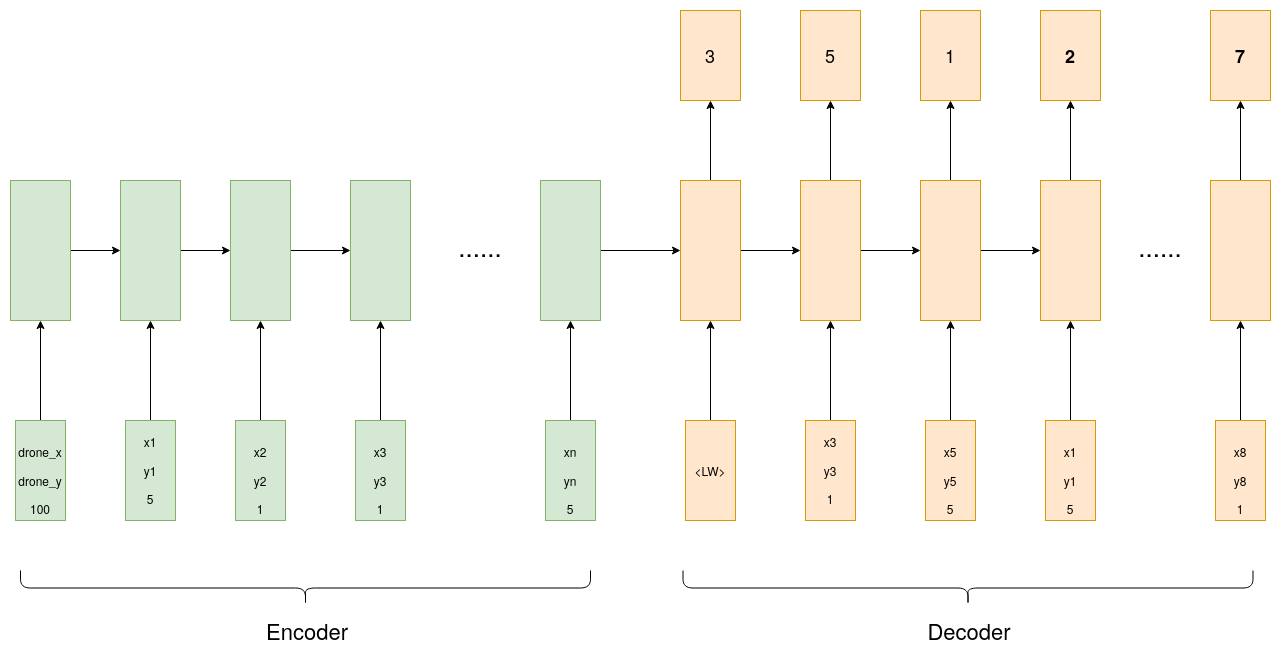}
    \caption{The Seq2Seq model has an encoder-decoder architecture with LSTM cells. This image shows a single layer LSTM encoder-decoder model, our experiments use 2 layers. $< LW >$ represents a  learnable parameter}
    \label{fig:seq2seq}
\end{figure}

The architecture of the Seq2Seq model \cite{sutskever_sequence_2014} used in our experiments is shown in Figure \ref{fig:seq2seq}. The input to the encoder part of the model consists of the drone location, location of all of the PoI and its priority level. The encoder block of the model compresses the input information into a fixed dimensional vector called context vector. The decoder block uses the context vector and predicts the location the drone has to visit at each time step. The output at the previous time step is provided as input to the current time step for the decoder block. At the first time step, the input to the decoder block is a learnable vector $<LW>$.

\subsubsection{Seq2Seq with Attention}
\begin{figure}
    \centering
    \includegraphics[width=0.8\linewidth]{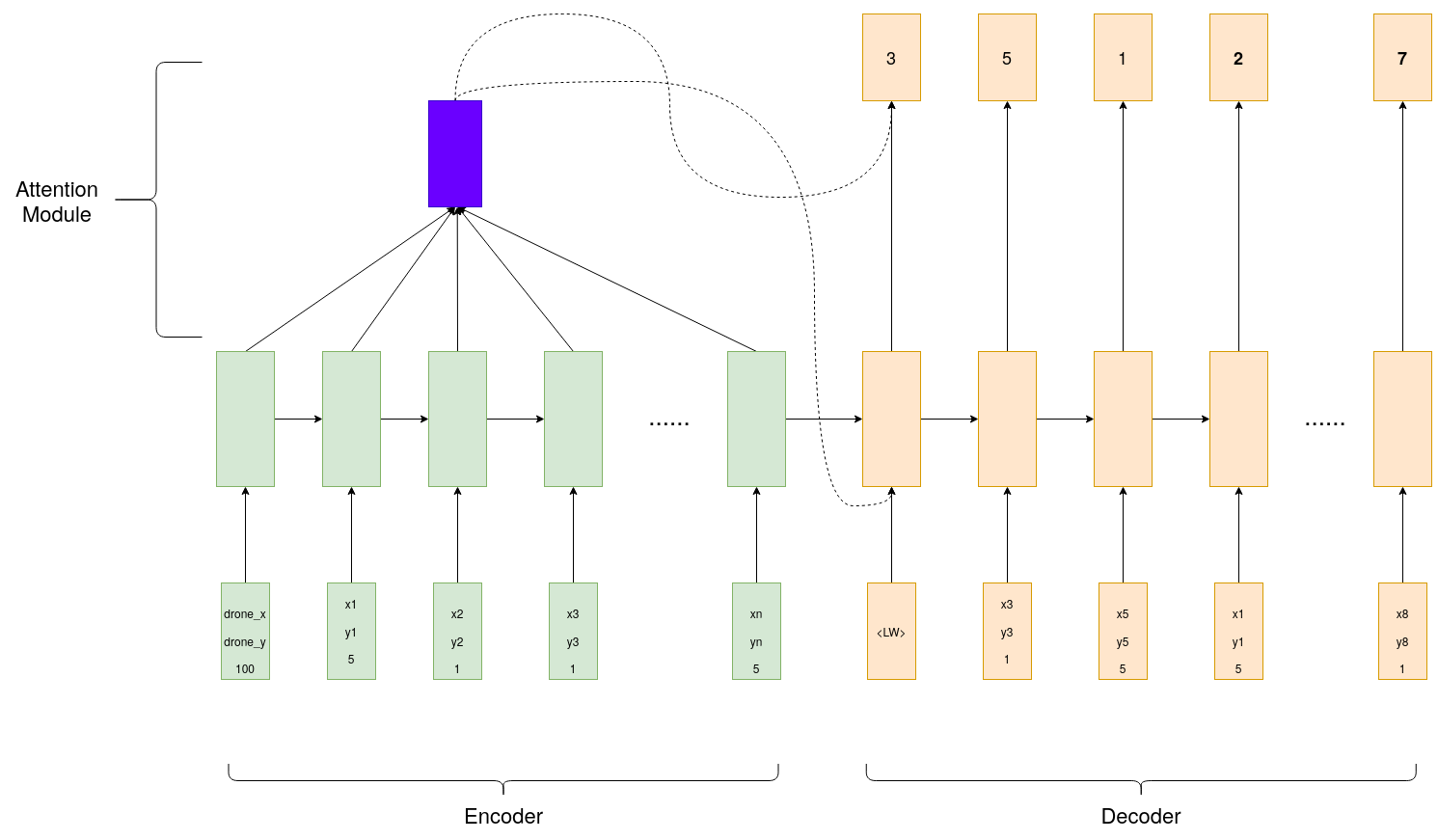}
    \caption{The Seq2Seq model with attention has an encoder-decoder architecture with LSTM cells. The decoder at each time step receives additional context from an attention module. $<LW>$ represents a learnable parameter.}
    \label{fig:seq2seqAttn}
\end{figure}

The architecture of the Seq2Seq model with attention \cite{bahdanau_neural_2016} is shown in Figure \ref{fig:seq2seqAttn}. The decoder at each step receives additional context from the attention module which is a weighted combination of the hidden states of the encoder model. The input to the encoder and decoder blocks of the model are identical to that of the Seq2Seq model.

\subsubsection{Transformers}
\begin{figure}
    \centering
    \includegraphics[width=0.9\linewidth]{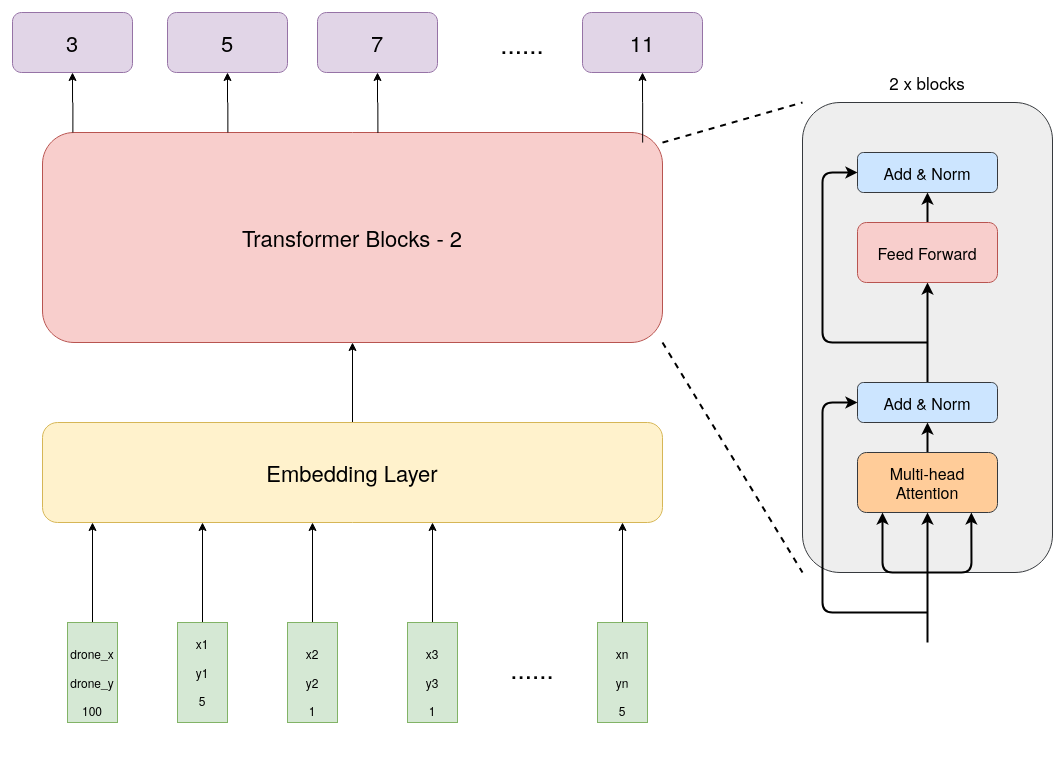}
    \caption{The input contains location of PoI and its priority level which is then
    projected onto a higher dimensional space using an embedding layer. The embeddings are
    then provided as input to the transformer encoder blocks.}
    \label{fig:transformer}
\end{figure}

The architecture of the transformer \cite{vaswani_attention_2017} model is shown in Figure \ref{fig:transformer}. The input of the model contains drone location, location of PoI and its priority level. This information is projected into a high dimensional space through an embedding layer which is a non-linear projection of the input. The high dimensional vectors are then passed through the two Transformer blocks. Each transformer block consists of a multi-head attention module, layer norm \cite{ba_layer_2016} and a feed forward layer with skip connections. The output of the transformer model is projected back to a lower-dimensional space to predict the point to visit at each time step.

\subsubsection{Pointer Net} \label{ptr_net}
\begin{figure}
    \centering
    \includegraphics[width=0.8\linewidth]{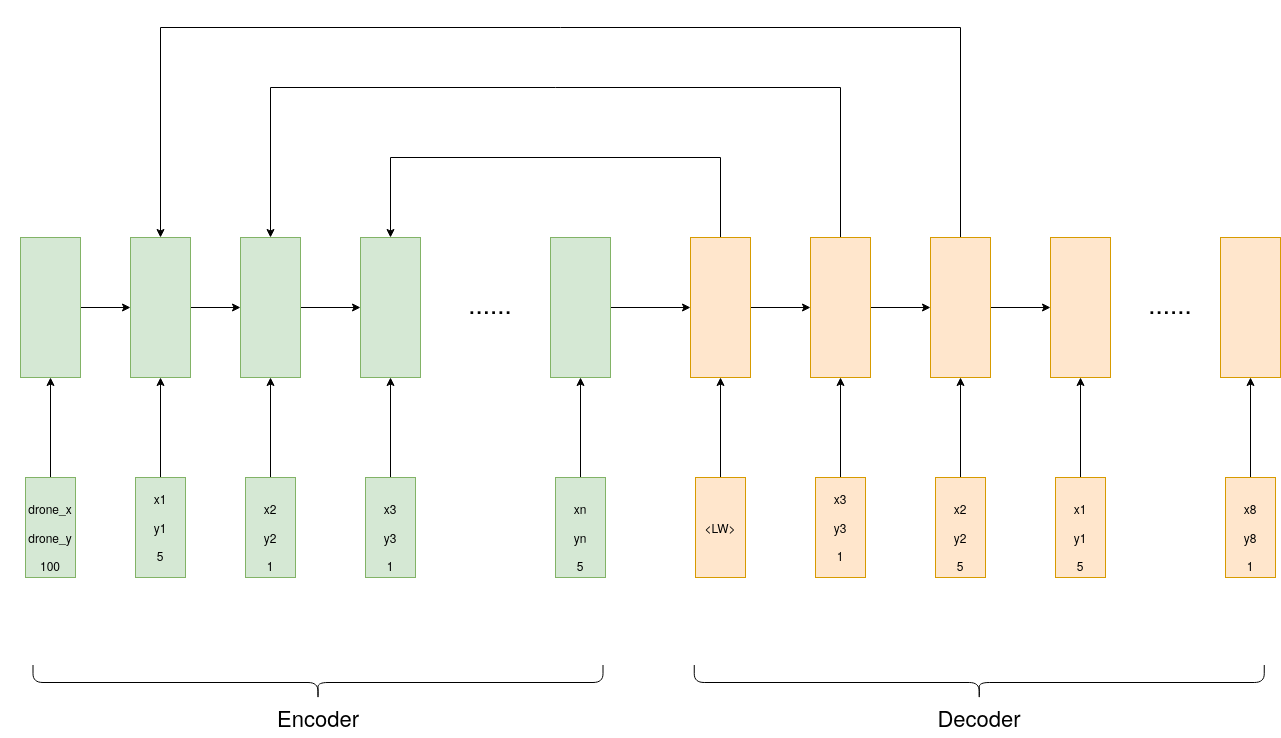}
    \caption{The PointerNet has an encoder-decoder architecture with LSTM cells.
    The decoder at each step points to the location in within the encoder states that the drones have to visit. $<LW>$ represents a learnable parameter.}
    \label{fig:pointernet}
\end{figure}

The architecture of the Pointer Net \cite{vinyals_pointer_2017} is shown in Figure \ref{fig:pointernet}. The input to the encoder and decoder blocks of the model are identical to that of the Seq2Seq model. At each step, the decoder points to location in the encoder block that it should visit.

\subsection{Results}
The results of the different policy networks are shown in Table \ref{tab:pol_result}. The time steps reported in the table depicts the number of time steps a single agent takes to map a $5\times 5$ grid. We were able to observe that the Seq2Seq model takes the most number of steps compared to other policy networks. This is reasonable as all other models use additional context in the form of attention modules to make the prediction. Therefore, we can infer that the attention modules help in improving the prediction. However, the Seq2Seq model has the least number of parameters among all of the mentioned policy networks. Therefore, there is a trade-off to be made between performance (number of time steps) and computation. 

\begin{table}[h]
\centering
\begin{tabular*}{\columnwidth}{cccc}
\hline
Policy Network & Parameters & CPU Wall time (ms) & Time steps \\
\hline
Seq2Seq \cite{sutskever_sequence_2014} & 103516 & 15.6 & 60 \\
Seq2Seq-Attention \cite{bahdanau_neural_2016} & 161564 & 22.3 & 55 \\
Transformer \cite{vaswani_attention_2017} & 111140 & 16.4 & 55 \\
Pointer Net \cite{vinyals_pointer_2017} & 116736 & 31.2 & 56 \\
\hline
\end{tabular*}
\caption{Comparison of different policy networks based on time steps a single drone/agent takes to map $5\times 5$ grid, CPU wall time on Intel Core i7 with 6 cores and number of parameters in the network. We assume the time taken for performing each action is equal, therefore, total time steps and total steps are equivalent. Note that CPU wall time is an unreliable metric as it is dependent on the background processes and CPU performance}
\label{tab:pol_result}
\end{table}

\subsection{Discussion}
\subsubsection{Plug and Play transfer learning}
An advantage of using these models is the ability to scale-up to multiple agents/drones without additional training, i.e. we can train a single agent using the REINFORCE algorithm and replicate the parameters of the model to multiple agents and perform inference. This allows us to save computational resources during the training phase.\\

\subsubsection{Disadvantages}
The disadvantage of using the above-mentioned policy networks is that the order in which the PoI are visited (policy prediction) is done at the beginning. In real life, however, the drones might have drifted off the desirable path due to sensory measurement errors or to navigate around an obstacle. To adjust the predicted action for the correction, we must perform another forward pass through the policy network which would increase the computational requirements.

\section{Path Planning}
\begin{figure}
    \centering
    \includegraphics[width=0.8\linewidth]{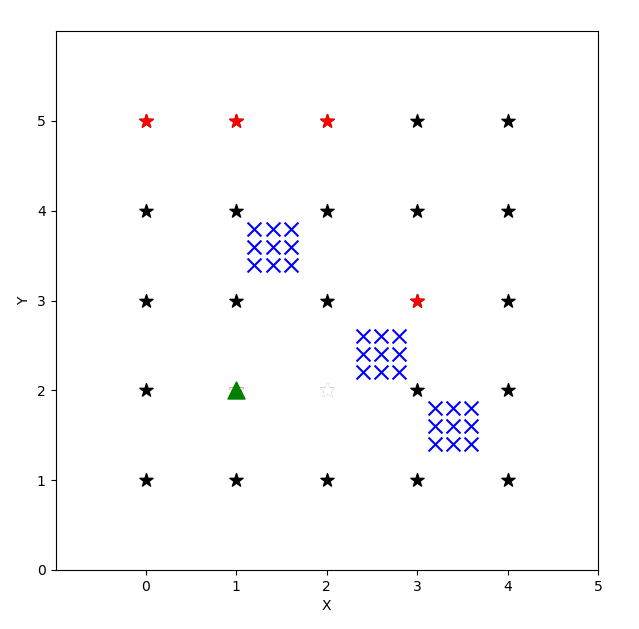}
    \caption{A rendered image from the 2D simulation of a 5x5 grid with obstacles. The green triangle represent the drones, the stars represent the PoI. The red stars have a higher priority level compared to black stars. The blue X's are obstacles present in the environment}
    \label{fig:obs_sim}
\end{figure}

The policy network predicts the order in which the drones have to visit the PoI to ensure fast coverage of all points. In real life, however, moving between one point of interest to another is not straightforward as the drones might have to avoid obstacles on their paths to reach the goals. We take the prediction of our policy network and perform path planning to avoid collision with obstacles. We compare two path planning algorithms for obstacle avoidance: Wavefront planner and Potential Field planner, which are tested in 2D and 3D simulations with obstacles present in the environment. A rendered image of the 2D simulation with obstacles is shown in Figure \ref{fig:obs_sim}. Note that the Pointer Net described in the section \ref{ptr_net} is used to generate the order in which the PoI have to be visited by the drones and the path planning algorithms only performs obstacle avoidance to reach the goal state, i.e. the assigned point at that time step for each drone. 

\subsection{Wavefront Planner}
The Wavefront planning algorithm \cite{Zelinsky93planningpaths} is a grid-based global planning algorithm, i.e. it assumes that the entire environment is visible to drones and the obstacles present in the environment are static. Any changes to the location of the obstacle have to be updated to the drones and the planning algorithm has to be re-run to make adjustments to the trajectory. The Wavefront trajectory planning algorithm works as follows:

\begin{itemize}
    \item The environment is split into discrete grids and the goal cell is initialized with value 2, obstacle cells with value 1 and all other cells with value 0
    \item At iteration 1, start from the goal state and choose all reachable cells based on connectivity and initialize them with value 2 + 1 = 3
    \item At iteration 2, choose all reachable cells from cells with value 3 and initialize them with value 3 + 1 = 4
    \item Perform the above step iteratively until you reach the drone's location cell. Do not consider obstacle cells with value 1 while performing the above steps
    \item To obtain the shortest path from the drone’s location to the goal state, choose the cell
    with the lowest value at each time step.
\end{itemize}

The algorithm performs a breadth-first search among the grids to predict the shortest path between the drone and the goal state. The disadvantage with the Wavefront planning algorithm is the assumption that the obstacles have to be static and the entire environment space is always known to the drones. Another disadvantage is the discrete nature of the algorithm. In real life, the drones can only move in continuous velocities and distances and cannot make discrete jumps. Increasing the granularity of the algorithm, i.e. increasing the number of grid cells can mitigate this issue to a certain extent, however, this increase comes at a higher computational cost since we would have to perform a breadth-first search through a larger number of cells.

\subsection{Potential Field Planner - 2D Simulation} \label{pf_2d}
The Potential Field \cite{1087247} planning algorithm is a gradient-based local planning algorithm, i.e. it does not require global information about the environment and the objects only within the \emph{Field of View (FoV)} of the drone have an impact on its trajectory. The Potential Field planning algorithm minimizes a cost function characterized by attractive potential between the goal state and drone and a repulsive potential between the drone and the obstacles within its FoV. Note that other drones within the FoV of a particular drone are also considered as obstacles. We minimize the cost function with respect to the position of the drone and update the position by calculating the gradients to the cost function. This would provide us with a trajectory to the goal state.

\begin{equation} \label{pot}
    C = U_{att} + U_{rep}
\end{equation}

\begin{equation}
    L_i(x,y) = 
    \begin{cases} 
      0.5(x_i - y_i)^2 & \text{if} \mid x_i - y_i \mid < \delta \\
      \delta(\mid x_i - y_i \mid - 0.5\delta) & \text{otherwise}
      \end{cases}
\end{equation}

\begin{equation}
    U_{att} = L_i(x, g)
\end{equation}

\begin{equation}
    U_{rep} = 
    \begin{cases} 
      -L_i(x, o) & \text{if dist} \le \text{D} \\
      0 & \text{otherwise}
      \end{cases}
\end{equation}

\begin{figure}[h]
    \centering
    \includegraphics[width=0.9\linewidth]{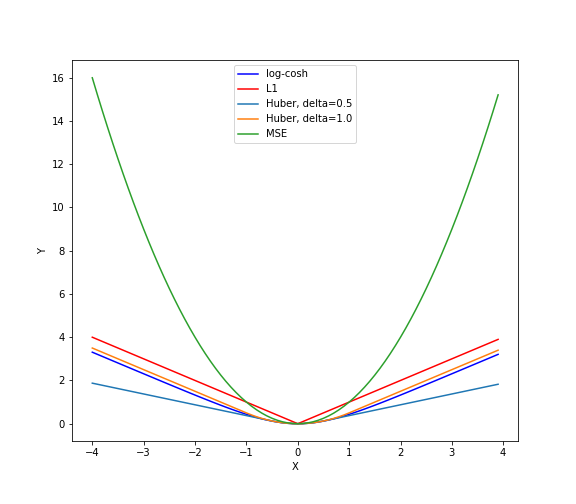}
    \caption{Visualization of $L_1$ loss, MSE, Huber loss and Log-Cosh functions}
    \label{fig:loss_func}
\end{figure}

The cost function used to model the Potential Field algorithm that we aim to minimize is shown in Equation (\ref{pot}). Huber loss, $L_1$ loss and Mean-Squared-Error (MSE) are among the most commonly used loss functions for regression tasks. In our task of path planning, these loss functions help minimize the distance between the drone and the goal state while maximizing the distance between the drone and the obstacles. They also have an implication in the trajectory of the drone. The different loss functions explored in our work are visualized in Figure \ref{fig:loss_func}.

\subsubsection{$L_1$ loss}
The $L_1$ loss is represented in Equation (\ref{l1}). This loss is less sensitive to outlier values compared to MSE as we are only calculating the absolute difference between the values. Therefore, it is a linear equation which provides only constant gradients to update the position of the drone irrespective of its distance to the goal or obstacle. The drawback of this function is that the derivative does not exist at value 0. To overcome this issue, we consider a point to be mapped when the Euclidean distance between the drone and the point of interest is below a threshold value $\gamma$. In our experiments, we set $\gamma$ to $0.5$.  

\begin{equation} \label{l1}
    L_i(x,y) = \mid x_i - y_i \mid
\end{equation}

\subsubsection{MSE loss}
The Mean-Squared-Error loss is represented in Equation (\ref{mse}). Relative to $L_1$ loss, MSE is sensitive to outliers in the data, as it is a quadratic representation of the error value. The loss function provides linear gradients, however, the gradients exhibit high values compared to $L_1$ loss when the goal point is located at a significant distance from the drone. 

\begin{equation} \label{mse}
    L_i(x,y) = (x_i - y_i)^2
\end{equation}

\subsubsection{Log-Cosh loss}
The log-cosh function is the logarithm of the hyperbolic cosine function. Equation (\ref{logcosh}) describes the loss function. For very high values of error, the $e^x$ term dominates in Equation (\ref{cosh}) and for very low values of error, $e^{-x}$ term dominates. Therefore the loss function behaves as a linear function at extreme values. At error values close to zero, the function can be approximated to a quadratic function using the \emph{Taylor series} expansion (Refer to Appendix for proof). 
 
\begin{equation} \label{logcosh}
    L_i(x,y) = \log(\cosh(x_i - y_i))
\end{equation}

\begin{equation} \label{cosh}
    \cosh(x) = \frac{e^x + e^{-x}}{2}
\end{equation}

\subsubsection{Huber Loss}
Huber loss \cite{huber1964} function is represented in Equation (\ref{huber}). The loss function uses a thresholding value $\delta$, where if the $L_1$ loss is below that value, a scaled value of the Mean-Squared-Error is returned as the loss value, else, the $L_1$ loss scaled by $\delta$ is the loss value where $\delta$ governs the slope of the loss function. A high value of $\delta$ will provide relatively higher gradients compared to a low value of $\delta$. 


\begin{equation} \label{huber}
    L_i(x,y) = 
    \begin{cases} 
      0.5(x_i - y_i)^2 & \text{if} \mid x_i - y_i \mid < \delta \\
      \delta(\mid x_i - y_i \mid - 0.5\delta) & \text{otherwise}
      \end{cases}
\end{equation}

\subsubsection{Adaptively Choosing $\delta$ - 2D} \label{adapt_delta_2d}
\begin{figure}[h]
    \centering
    \includegraphics[width=0.8\linewidth]{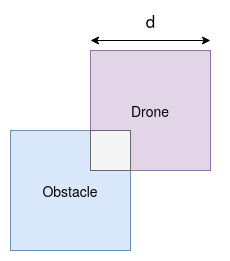}
    \caption{Area of FoV is represented by the violet square, the blue colored square represents the actual size of the obstacle and the grey square represents perceived size of the obstacle by the agent, i.e.  ($\text{Area of FoV} \cap \text{Area of Obstacle}$).}
    \label{fig:ada_delta}
\end{figure}

We explore different methods to adaptively choose $\delta$ which governs the magnitude in which the drone moves at each time step. A higher value of $\delta$ would result in larger steps in the environment or a larger velocity (in terms of step size per time step) which causes the agent to collide with an obstacle which is not in its FoV. Choosing a low value of $\delta$ would result in the agent being more cautious in terms of navigation, i.e. making smaller updates. This, however, might not be optimal because the agent might end up taking smaller steps even though there are no obstacles in its FoV. Using this intuition, we model the equation for choosing $\delta$ based on the Equations (\ref{adapt_2d_scale}); an illustration is provided in Figure \ref{fig:ada_delta}. The Equation (\ref{adapt_2d_scale}) with parameters $\alpha=1$ and $\beta=1$, $\alpha=2$ and $\beta=1$, $\alpha=4$ and $\beta=1$ can be grouped together as the perceived size of the obstacle by the drone is \emph{linearly scaled} and with parameters $\alpha=1$ and $\beta=2$, the perceived size of the obstacle is \emph{quadratically scaled}, $d$ represents the length of the edge of the FoV, shown in Figure \ref{fig:ada_delta}. 

\begin{equation} \label{adapt_2d_scale}
    \delta = \frac{\text{Area of FoV}}{\text{Area of FoV} + \alpha(\text{Area of FoV} \cap \text{Area of Obstacle})^{\beta}} \times d/2
\end{equation}





\hfill

The area of FoV is the area that is visible to the agent and Area of an obstacle is the perceived size of the obstacle within the agent's FoV. Note that the calculation of $\delta$ is independent of the modality of the perception, however, dependent on its accuracy in estimating the perceived size of the obstacle. Incorrect or noisy estimates might have negative consequences
as $\delta$ could be set to a high value which could cause a collision with an obstacle.

\begin{figure}[h]
    \centering
    \includegraphics[width=0.9\linewidth]{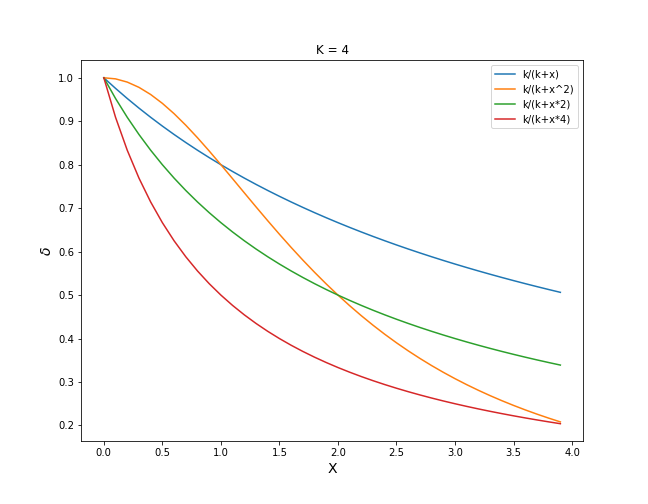}
    \caption{Curves representing Equation (\ref{adapt_2d_scale}) with parameters $\alpha=1$ and $\beta=1$, $\alpha=2$ and $\beta=1$, $\alpha=4$ and $\beta=1$, $\alpha=1$ and $\beta=2$. The x-axis represents the perceived size of the obstacle, i.e. $\text{Area of FoV} \cap \text{Area of Obstacle}$ and y-axis represents the value of $\delta$. K represents the area of the FoV which is 4 units in our example.}
    \label{fig:k_delta_graph}
\end{figure}

Figure \ref{fig:k_delta_graph} shows how the value of $\delta$ varies based on the perceived size of the obstacle for different equations. When using Equation (\ref{adapt_2d_scale}) with parameters $\alpha=4$ and $\beta=1$ as the method for choosing $\delta$, we were able to observe that sometimes when the obstacle is close to the agent, the gradient update becomes small as $\delta$ is aggressively decreased, which causes the drone to take very small steps. In order to encourage the drones to take larger steps, we add small noise to the gradients, when the gradient update is below a threshold value.

\subsubsection{Results}
Table \ref{tab:path_plan_tab} shows the number of time steps 2 drones take to map a $5\times 5$ environment using the Potential Field planner. The number of time steps a single agent takes to map 25 points represented as $5\times 5$ grid using the Wavefront planner is 113. We do not report this score in Table \ref{tab:path_plan_tab} as it is not an appropriate comparison. 

\begin{table}[h]
\centering
\begin{tabular}{cccc}
\hline
Loss function & $\alpha$ & $\beta$ & Time steps \\
\hline
$L_1$ loss & - & - & 137 \\
MSE loss & - & - &228 \\
Log-cosh loss & - & - &122 \\
Huber loss - $\delta = 0.5$ & - & - &221 \\
Huber loss - $\delta = 1.0$ & - & - &94 \\
Adaptively choose $\delta$ - 2D & $1$ & $1$ &121 \\
Adaptively choose $\delta$ - 2D & $2$ & $1$ &138 \\
Adaptively choose $\delta$ - 2D & $4$ & $1$ &185 \\
Adaptively choose $\delta$ - 2D & $1$ & $2$ &173 \\
\hline
\end{tabular}
\caption{Comparison of number of time steps each loss function takes to map 25 points.  Time steps is measure of number of steps 2 drones take to map all points using the Potential Field planner in a 2D simulation. The cells are represented as a $5\times 5$ grid. We assume the time taken for performing each action is equal, therefore, total time steps and total steps are equivalent. The time steps reported are the best score across three runs. Note that the drones can move in continuous space, i.e. it could move to any real number coordinate within the environment.}
\label{tab:path_plan_tab}
\end{table}

\subsubsection{Discussion}
We observe that Mean-Squared-Error loss takes the most number of time steps to map an environment. This is because the gradient of the loss function is linearly scaled. This causes the drone to make large jumps when the distance between goal and drone is high and aggressively reduce distance covered when the drone approaches the goal. Huber loss function with $\delta$ set to 0.5 also takes a long time because, the low value of $\delta$ makes the drone more cautious in terms of navigation. Therefore, only small updates are made to the position of the drone, which leads to increased mapping time. Adaptively choosing $\delta$ through Equation (\ref{adapt_2d_scale}) with parameters $\alpha=4$ and $\beta=1$, $\alpha=1$ and $\beta=2$ also increases the mapping time for a similar reason, i.e. the presence of a small obstacle within the FoV aggressively reduces $\delta$ compared to using parameters $\alpha=1$ and $\beta=1$, $\alpha=2$ and $\beta=1$. This causes small updates to the position of the drone which leads to an increase in time steps. 

\subsection{Potential Field Planner - 3D simulation}
\begin{figure}
    \centering
    \includegraphics[width=\linewidth]{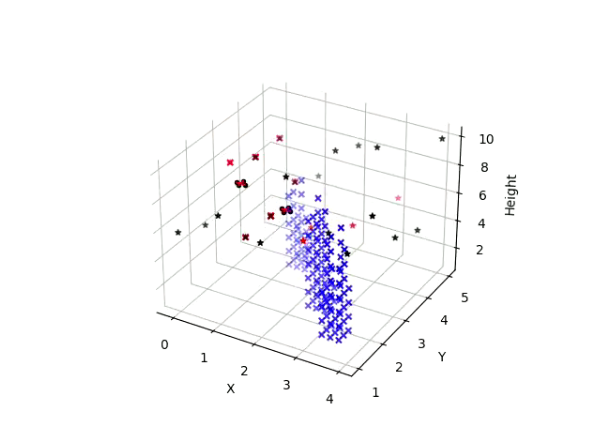}
    \caption{A screenshot from the 3D simulation. The black and red stars represent PoI with low and higher priority respectively and are placed at random altitudes from range $[5, 10]$ units within the x-y grid. The blue "X" represent the obstacles and the height of each obstacle is randomly assigned a value from range $[1, 10]$ units. A red "X" is drawn over a cell when it is mapped by a drone.}
    \label{fig:pf_3d}
\end{figure}

We use the Potential Field planner in a 3D simulation and discuss the effects of different loss functions on the trajectory of the drone. In the 3D simulation, we dynamically vary the altitude of the obstacles and the Points of the Interest to be mapped. Figure \ref{fig:pf_3d} represents a screenshot from the simulation with obstacles of random altitude and PoI present at different heights. 

A point is considered to be mapped when the $L_2$ loss between the drone's center and point's center is below a threshold value. We observe that in the case of varying both the obstacle height and PoI altitude, the drones oscillate when there is an obstacle within its FoV. To nudge the drone out of this oscillation, we reuse the earlier trick of adding small noise to the gradient. An ideal behaviour would be that, when a drone is close to an obstacle, it would move tangentially to the obstacle allowing it to move closer to the goal while avoiding collision. Adding small noise to the gradient allows us to achieve this but modifying gradients could translate into small jitters in the motion of the drone which is not desirable and also a waste of power resource. Therefore, we scaled the repulsive potential in the cost function by a value from range $(0, 1)$. Scaling the contribution of repulsive potential to the cost function reduces its effect but does not eliminate it. Instead of a direct step back, the attractive potential contributes relatively more to the gradient of the cost function, therefore, causing it to move tangentially to an obstacle and breaking the oscillation. 

\subsubsection{Adaptively Choosing $\delta$ - 3D}
\begin{figure}[h]
    \centering
    \includegraphics[width=0.8\linewidth]{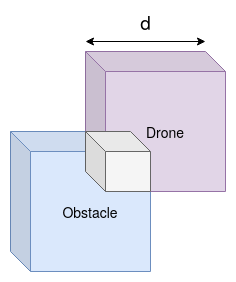}
    \caption{Volume of FoV is represented by the violet cube, blue colored cube represents the actual size of the obstacle and the grey cube represents perceived size of the obstacle by the agent, i.e. ($\text{Vol. of FoV} \cap \text{Vol. of Obstacle}$).}
    \label{fig:3d_ada_delta}
\end{figure}

We also evaluate the Huber loss function in 3D for path planning of the drones. We reuse and adapt the equations provided in Section \ref{adapt_delta_2d} for the 3D simulation. The Equations are shown below. We consider the volume represented by the FoV instead of the area and we use intersection of the volumes between the perceived size of the obstacle and the FoV of the drone to scale the value of $\delta$. The value $d$ in the equation represents the length of the edge of the cube representing the FoV of the drone. This is illustrated in Figure \ref{fig:3d_ada_delta}.

\begin{equation} \label{adapt_3d_scale}
    \delta = \frac{\text{Vol. of FoV}}{\text{Vol. of FoV} + \alpha(\text{Vol. of FoV} \cap \text{Vol. of Obstacle})^{\beta}} \times d/2
\end{equation}





\begin{figure}[h]
    \centering
    \includegraphics[width=0.9\linewidth]{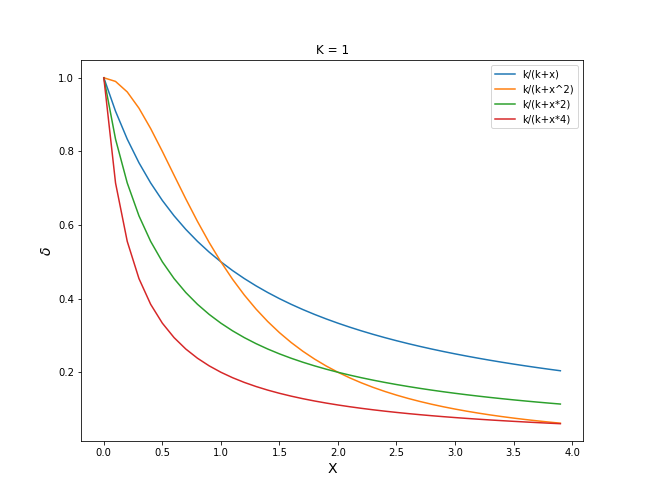}
    \caption{Curves representing Equation (\ref{adapt_3d_scale}) with parameters $\alpha=1$ and $\beta=1$, $\alpha=2$ and $\beta=1$, $\alpha=4$ and $\beta=1$, $\alpha=1$ and $\beta=2$. The x-axis represents the perceived size of the obstacle, i.e. $\text{Vol. of FoV} \cap \text{Vol. of Obstacle}$ and y-axis represents the value of $\delta$. K represents the volume of the FoV which is 1 unit in our example.}
    \label{fig:k_1_delta_graph}
\end{figure}

\hfill \break
Figure \ref{fig:k_1_delta_graph} shows how the value of $\delta$ scales based on the intersection of volumes. We were able to observe that Equation (\ref{adapt_3d_scale}) with parameters $\alpha=4$ and $\beta=1$ reduces $\delta$ aggressively than other parameter choices. This translates into very smaller steps taken by the drone in the simulation when there is an obstacle within its FoV. Therefore, the drones take a longer time to map all the PoI.

\subsubsection{Results}
The $L_1$ loss function is used as the baseline for comparing performance in the 3D simulation. Time steps are a measure of number of total steps required to map all points in the environment. The results reported in Table \ref{tab:pf_result_3d} are for 2 drones and for 25 points present in the environment at random altitudes. The obstacles are also of varying height. 


\begin{table}[h]
\centering
\begin{tabular}{cccc}
\hline
Loss function & $\alpha$ & $\beta$ & Time steps \\
\hline
$L_1$ loss & - & - & 561 \\
Adaptively choose $\delta$ - 3D & $1$ & $1$ & 309 \\
Adaptively choose $\delta$ - 3D & $2$ & $1$ & 371 \\
Adaptively choose $\delta$ - 3D & $4$ & $1$ & 469 \\
Adaptively choose $\delta$ - 3D & $1$ & $2$ & 524 \\
\hline
\end{tabular}
\caption{Comparison of number of time steps for $L_1$ loss and Huber loss function with adaptively choosing $\delta$. Time steps is measure of number of steps 2 drones take to map 25 points using the Potential Field planner in a 3D simulation. We assume the time taken for performing each action is equal, therefore, total time steps and total steps are equivalent. The cells are represented as a $5\times 5$ grid. The time steps reported are the best score across three runs. Note that the drones can move in continuous space, i.e. it could move to any real number coordinate within the environment.}
\label{tab:pf_result_3d}
\end{table}

\subsubsection{Discussion}
The number of time steps taken when using the $L_1$ loss function for the Potential Field algorithm is relatively higher compared to methods of adaptively choosing $\delta$. We could observe that adaptively changing $\delta$ based on parameters $\alpha=4$ and $\beta=1$, $\alpha=1$ and $\beta=2$ increases the time steps. This is because they aggressively decrease $\delta$ compared to parameters $\alpha=1$ and $\beta=1$, $\alpha=2$ and $\beta=2$. Decreasing $\delta$ causes the drone to make smaller updates to its position. We could also observe that parameter $\alpha=1$ and $\beta=1$ maps the environment faster than other functions. 

\section{Conclusion \& Future Work}
We have performed a comparative analysis on several neural network architectures using RL for performing agent-oriented task assignment based on their effectiveness and efficiency. We couple the task assignment with path planning algorithms and perform obstacle avoidance without global knowledge of the obstacles in the environment and plan locally, i.e. plan actions at the edge. We also proposed a method to adaptively change velocity to allow safety during obstacle avoidance. These evaluations and results would ease the decision of choosing methods for performing task assignment and path planning for robots in a multi-agent setting, not restricted to drone swarms.  

This work assumes that the task assignment and the path planning problems are disjoint. These problems are solved independently by producing approximate solutions. Therefore, in certain cases (e.g.: an environment with a very large number of obstacles) this leads to poor approximation and hence, a bad solution. This could be alleviated if we could conflate the task assignment and the path planning task under a single/universal framework where the agent learns to reach goal states and avoids obstacles in tandem. We hypothesize that using Graph Neural Network as a policy network and RL algorithms for training could allow us to achieve this. We seek to explore this in future work. 


\section*{Acknowledgment}
This research was supported by RiskEcon\textsuperscript{®} Lab for Decision Metrics @ Courant Institute of Mathematical Sciences NYU and Agile Robotics and Perception Lab (ARPL) @ NYU Tandon School of Engineering with funding provided by Numerati\textsuperscript{®} Partners. 


\bibliographystyle{IEEEtran}
\bibliography{main}

\section*{Appendix} \label{Appendix}
The Taylor series expansion of $e^x$ and $e^{-x}$ are given in the two equations below.
\begin{equation}
    e^x = 1 + x + \frac{x^2}{2!} + \frac{x^3}{3!} + \ldots
\end{equation}

\begin{equation}
    e^{-x} = 1 - x + \frac{x^2}{2!} - \frac{x^3}{3!} + \ldots
\end{equation}

\begin{align*}
    \cosh(x) &= \frac{e^x + e^{-x}}{2} \\
    \implies \frac{e^x + e^{-x}}{2} &= 1 + \frac{x^2}{2!} + \frac{x^4}{4!} + \ldots \\
    \therefore \log{\cosh(x)} &= \log(1 + \frac{x^2}{2!} + \frac{x^4}{4!} + \ldots)
\end{align*}

\begin{align*}
    \text{Let } \alpha &= \frac{x^2}{2!} + \frac{x^4}{4!} + \ldots \\
    \implies \log{\cosh(x)} &= \log(1 + \alpha)
\end{align*}

Using the Taylor series expansion of $\log(1+x)$ we approximate the above equation.

\begin{align*}
    \log(1 + \alpha) &= \alpha - \frac{\alpha^2}{2} + \frac{\alpha^3}{3} + \ldots \\
    \text{By substituting for } \alpha \\
    \therefore \log{\cosh(x)} &= \frac{x^2}{2!} + \frac{x^4}{4!} + \ldots + {(\frac{x^2}{2!} + \frac{x^4}{4!} + \ldots)}^2 + \ldots
\end{align*}

The above equation can be approximated to $x^2$ when $-1<x<1$.

\begin{align*}
    \therefore \log{\cosh(x)} \approx x^2 \text{ when } -1<x<1
\end{align*}

\end{document}